\documentclass[sn-mathphys]{sn-jnl}

\usepackage{amsmath}
\usepackage{amssymb}
\usepackage{booktabs}
\usepackage{multirow}
\usepackage{bm}
\usepackage{commath}

\theoremstyle{thmstyleone}%
\theoremstyle{thmstyletwo}%

\theoremstyle{thmstylethree}%

\begin{document}

\title[End-to-End Autoencoder for Drill String Acoustic Communications]{End-to-End Autoencoder for Drill String Acoustic Communications}


\author*[1]{\fnm{Iurii} \sur{Lezhenin}}\email{lezhenin.yu@edu.spbstu.ru}

\author[1]{\fnm{Aleksandr} \sur{Sidnev}}\email{sidnev\_ag@spbstu.ru}

\author[1]{\fnm{Vladimir} \sur{Tsygan}}\email{tsygan\_vn@spbstu.ru}

\author[1]{\fnm{Igor} \sur{Malyshev}}\email{malyshev\_ia@spbstu.ru}

\affil*[1]{\orgdiv{Institute of Computer Science and Cybersecurity}, \orgname{Peter the Great St. Petersburg Polytechnic University}, \orgaddress{\street{Polytechnicheskaya~29}, \city{St.Petersburg}, \postcode{195251}, \country{Russia}}}



\abstract{Drill string communications are important for drilling efficiency and safety. 
The design of a low latency drill string communication system with high throughput and reliability remains an open challenge.
In this paper a deep learning autoencoder (AE) based end-to-end communication system, where transmitter and receiver implemented as feed forward neural networks, is proposed for acoustic drill string communications.
Simulation shows that the AE system is able to outperform a baseline non-contiguous OFDM system in terms of BER and PAPR, operating with lower latency.}




\maketitle

\section{Introduction}\label{sec1}

During drilling a borehole for oil or gas extraction, a large amount of data, acquired from the drilling rig sensors, has to be transmitted to the surface in real-time. This process is known as measurement while drilling (MWD). 
Nowadays, the most promising method for high speed MWD communications is acoustic drill string transmission \cite{Gao2008}. 


Drill string acoustic channel has frequency-selective fading and a long impulse response \cite{Wang2007, Han2013}. 
Orthogonal frequency-division multiplexing (OFDM) is widely used for MWD communications \cite{Gutierrez-Estevez2013, Ma2017, Ma2018}, as it is robust to frequency-selective fading. 
OFDM based MWD systems provide good throughput and reliability, but have poor latency because of a long impulse response of the channel.
A cyclic prefix (CP) which is required to mitigate the intersymbol interference (ISI) should be as long as channel impulse response
and, consequently, to achieve the acceptable CP overhead, OFDM operates with symbols of long duration. %
For drill string acoustic communications a design of low latency system with respect to reliability and throughput requirements remains an open challenge.

Deep learning autoencoder (AE) based end-to-end communication systems are the current state-of-the-art, where transmitter and receiver are implemented by neural networks (NN). They are trained as an autoencoder to adapt to a specific communication channel. During training transmitter and receiver are optimized jointly for end-to-end performance. As a result autoencoder learns signal waveforms, which are resistant to channel impairments.



It was shown that AE systems are able to perform as good as uncoded quadrature amplitude modulation (QAM) and Hamming coded QAM modulation schemes under additive white Gaussian noise (AWGN) and flat Rayleigh fading channels \cite{OShea2017, Wu2019}. Under certain fading channels AE outperforms QAM scheme with minimum mean square error (MMSE) equalization \cite{Zhu2019_1}, and bi-directional recurrent NN based AE is comparable with QAM with maximum-likelihood sequence estimation (MLSE) equalizer \cite{Wu2020}.


In real-world applications, when channel model is known, but is too complicated to derive an optimal modulation scheme, AE is able to provide a better solution than conventional systems.
Much attention is payed to AE systems in context of optical communications \cite{Karanov2018, Karanov2019, Zhu2019} including those using nonlinear dispersive channels; an AE system was designed and applied for human body communications \cite{Ali2019}; AE is applicable for molecular communications \cite{Mohamed2019} as well. 

In this paper the perspective of applying AE for acoustic drill string communications is studied. An AE communication system is proposed and evaluated. Its performance in terms of bit error rate (BER) and signal peak-to-average power (PAPR) is compared with a baseline OFDM system. Simulation shows that for drill string communications AE may provide reliable transmission with lower latency at the same  throughput.

\section{Drill String Communication Channel}\label{sec:channel}

In acoustic MWD system a downhole transmitter generates mechanical longitudinal waves, which propagate along a drill string to a surface receiver.
In the conventional acoustic channel model of a drill string (e.g. \cite{Gao2008}) the transmitted signal $x(t)$ with the added downhole drill bit noise $n_d(t)$ flows through acoustic channel with an impulse response $h(t)$. Then the signal is affected by the surface noise $n_s(t)$ produced by drilling equipment. Thus, at the receiver side the signal $y(t)$ is given as
\begin{equation*}
    y(t) = x(t) * h(t) + n_d(t) * h(t) + n_s(t).
    \label{eq:model_time}
\end{equation*}

\subsection{Frequency Response of Drill String}

\begin{figure}[h]
    \centering
    \includegraphics[width=0.95\textwidth]{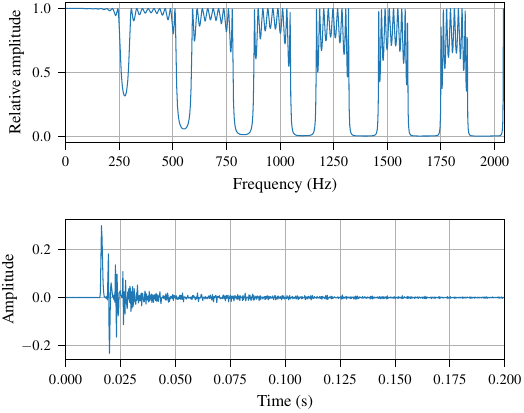}   
    \caption{A frequency response and impulse response obtained using transfer matrix method for a drill string composed of 10 pipes and 9 joints (Table \ref{tab:drillparams}).}
    \label{fig:resp}
\end{figure}

A drill string consists of alternating drill pipes and tool joints. Such structure results in multiple reflections of acoustic waves and, therefore, channel response (Fig. \ref{fig:resp}) is similar to a comb filter. A frequency response of a geometrically ideal drill string is obtained using transfer matrix method \cite{Wang2007}.

Suppose, a drill string consists of $N$ tube elements, $d_n$ and $a_n$ are the length and the crossectional area of the $n$-th element, respectively, for $n = 1 \dots N$. The mass density of the drill string is $\rho$, and the velocity of longitudinal waves is $c$. The longitudinal waves are projected on the right surface of the $N$-th element and arrive on the left surface of the first element. 

The potential function of the $n$-th element is expressed as
\begin{equation*}
    \Phi^{(n)}(x) = \Phi^{(n)}_I e^{jkx} + \Phi^{(n)}_R e^{-jkx},
\end{equation*}
where $\Phi^{(n)}_I$ and $\Phi^{(n)}_R$ are the amplitudes of potential functional of incident and reflected waves, respectively, and $k$ is the longitudinal wave number. 
Let the reflection coefficient $R$ and transmission coefficient $T$ be defined as
\begin{equation*}
    T = \Phi^{(1)}_I / \Phi^{(N)}_I, \; R = \Phi^{(N)}_R / \Phi^{(N)}_I.
\end{equation*}
For the whole drill string there is a matrix equation
\begin{equation*}
    \pmb{A}_N(0) \begin{bmatrix} 1 + R \\ 1 - R \end{bmatrix} =  \pmb{M} \pmb{A}_1(0)  \begin{bmatrix} T \\ T \end{bmatrix},
\end{equation*}
where $\pmb{A}_n$ is a $2 \times 2$ matrix function defined 
as
\begin{equation*}
    \pmb{A}_{n}(x) = \begin{bmatrix}
    -k\sin(kx) & j \, k \cos(kx) \\ \rho \, a_n \, c^2 k^2 \cos(kx) & j \rho \, a_n \, c^2 k^2 \sin(kx)
    \end{bmatrix},
\end{equation*}
and $\pmb{M}$ is a $2 \times 2$ matrix given by
\begin{equation*}
    \pmb{M} = \pmb{A}_N(d_N) \pmb{A}_N^{-1}(0) \pmb{A}_{N-1}(d_{N - 1}) \pmb{A}_{N-1}^{-1}(0) \dots \pmb{A}_1(d_1) \pmb{A}_1^{-1}(0).
\end{equation*}
This equation can be solved about $R$ and $T$ numerically for certain drill string.

The transmission coefficient $T$ can be seen as a function of the wave number $k$. The wave number is obtained as  
\begin{equation*}
    k = \frac{2 \pi f}{c},
\end{equation*}
where $f$ is the longitudinal wave frequency. Thus, there is a relationship between $T$ and $f$, which is the frequency response $H(f)$ of a drill string channel. The impulse response $h(t)$ of the channel is obtained by using inverse Fourier transform. 

\subsection{Drill Noise}

The channel model includes two kinds of noise: drill bit noise $n_d(t)$ and surface noise $n_s(t)$. The both can be understood as overall drill noise 
\begin{equation*}
    n(t) = n_d(t) * h(t) + n_s(t).
    \label{eq:drill_noise}
\end{equation*}
The overall noise is Gaussian and fairly stationary, but its spectrum is not flat. 
Since true spectral shape depends on the surface equipment and, therefore, is usually unknown, it is common to use white Gaussian noise assumption.

\section{Autoencoder Communication System}\label{sec3}

\begin{figure}[h!]
\centering
\includegraphics[width=0.9\textwidth]{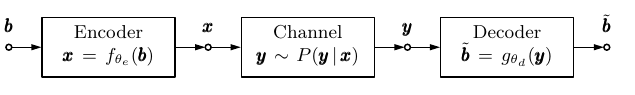}
\caption{Architecture of the autoencoder communication system.}
\label{fig:aearch}
\end{figure}

The AE communication system consists of three parts: the encoder, the decoder, and the channel (Fig. \ref{fig:aearch}).
Suppose, the sequence of $m$ bits $\pmb{b} \in \{0, 1\}^{m} $ is transmitted.
The encoder constructs the signal of $n$ complex samples $\pmb{x} \in \mathbb{C}^{n}$, an AE symbol, by mapping $f_{\theta_e} : \{0, 1\}^{m} \rightarrow \mathbb{C}^{n}$, where ${\theta_e}$ are the encoder trainable parameters. Hence, the communication rate of the system is $r = m / n$ (bits per channel use).
At the receiver side the noisy and distorted signal $ \pmb{y} \in \mathbb{C}^{n} $ is observed. The channel may be described by conditional distribution $ P(\pmb{y} \mid \pmb{x}) $. The decoder provides the estimation of original bit sequence $\tilde{\pmb{b}} \in \mathbb{R}^{m}$ by mapping $g_{\theta_d} : \mathbb{C}^{n} \rightarrow \mathbb{R}^{m}$, where ${\theta_d}$ are the decoder trainable parameters.
This estimation can be viewed as $\tilde{b}_i = P(b_i = 1 \mid \pmb{y})$, and $1 - \tilde{b}_i = P(b_i = 0 \mid \pmb{y})$, respectively. 


The average power constraint $\frac{1}{n}\sum_{i=1}^{n}|x_i|^2 = 1$ is imposed on the transmitted signal to make the implementation of such a system feasible.


\subsection{Encoder and Decoder}

The encoder and decoder are implemented as feedforward fully connected (dense) NNs (Table \ref{tab:pma}). The both consist of three layers. A rectified linear unit (ReLU) activation function is used to build an inner data representation, a linear activation allows to construct a signal of arbitrary form, a sigmoid maps logits to output probabilities. The proposed structure was optimized for the specific bit sequence length $m$ and symbol length $n$ (see Sec. \ref{sec:sim}). For larger values the structure should be increased to keep the performance.

\begin{table}[h]
    \caption{The structure of proposed autoencoder system.}
    \label{tab:pma}
    \begin{tabular}{r|rll}
    \toprule
    & Layer & Activation & Output Size \\
    \midrule
         \multirow{8}{*}{\rotatebox{90}{Encoder}} 
         & Input & & $m$ \\
         & Dense & ReLU & 512 \\
         & Batch Normalization & & 512 \\
         & Dense & ReLU & 1024 \\
         & Batch Normalization & & 1024 \\
         & Dense & Linear & $n \times 2$ \\ 
         & To Complex & & $n$ \\
         & Power Normalization & & $n$  \\
         \midrule 
         \multirow{4}{*}{\rotatebox{90}{Channel}}
         & Input & & $p \times n$ \\
         & Modulation & & $pun$ \\
         & Convoltion and AWGN & & $pun$ \\
         & Demodulation & & $p \times n$ \\
         \midrule 
         \multirow{7}{*}{\rotatebox{90}{Decoder}} 
         & Input & & $n$ \\
         & From Complex & & $n \times 2$ \\
         & Dense & ReLU & 1024 \\
         & Batch Normalization & & 1024 \\
         & Dense & ReLU & 512 \\
         & Batch Normalization & & 512 \\
         & Dense & Sigmoid & $m$ \\
         \bottomrule
    \end{tabular}
\end{table}

Autoencoder is trained to minimize the loss function
\begin{align*}
    \mathcal{L}(\theta_e, \theta_d) = \frac{1}{\abs{\mathcal{S}}}\sum_{\pmb{b} \in \abs{\mathcal{S}}} \bigg( \ell_{BCE}(\pmb{b}, \tilde{\pmb{b}}) + \alpha \, \ell_{PAPR} (\pmb{x}) \bigg),
\end{align*}
where $\mathcal{S}$ is the training set and $\mid\mathcal{S}\mid$ is the cardinality of $\mathcal{S}$. The term $\ell_{BCE}$ is the binary cross-entropy loss defined as  
\begin{align*}
    \ell_{BCE}(\pmb{b}, \tilde{\pmb{b}}) = - \frac{1}{m} \sum_{i=0}^{m} b_i \log(\tilde{b}_i) + (1 - b_i) \log( 1 - \tilde{b}_i).
\end{align*}
It measures the accuracy of bit sequence estimation.
The term $\ell_{PAPR}$ is the signal PAPR defined as  
\begin{align*}
    \ell_{PAPR}(\pmb{x}) = \frac{\max_{i = 1,\dots,n}{\abs{x_i}^2}}{\frac{1}{n}\sum_{i=1}^{n}\abs{x_i}^2} = \max_{i = 1,\dots,n}\abs{x_i}^2.
\end{align*}
Including PAPR into the loss function is the common approach of PAPR reduction for AE systems. 
In order to estimate true PAPR of continuous signal, in the above equation $\pmb{x}$ is upsampled by a factor of 4. The parameter $\alpha$ regulates the tradeoff between PAPR and transmission reliability. 

\subsection{Channel}    
A signal $\pmb{x}$ constructed by the encoder is baseband.
First, $\pmb{x}$ is upsampled by a factor $u$ and modulated by a carrier frequency $f_c$. Then a sequence of $p$ constructed independently signal frames are concatenated into a packet $\bar{\pmb{x}} \in \mathbb{C}^{pun}$. 
The whole packet is fed into the channel in order to simulate ISI distortions.
At the receiver side a distorted packet $\bar{\pmb{y}} \in \mathbb{C}^{pun}$ is split into separate signal frames. Then each frame is converted into baseband signal $\pmb{y}$ which is processed by the decoder.

The packet transmission through the channel is modeled as
\begin{equation*}
    \bar{\pmb{y}} = \bar{\pmb{x}} * \pmb{h} + \pmb{n},
\end{equation*}
where $\pmb{h} \in \mathbb{C}^{l}$ is a vector of $l$ discrete samples of the channel complex impulse response, and $\pmb{n} \in \mathbb{C}^{pun}$ is complex AWGN, i.e., $n_i \sim \mathcal{CN}(0, \sigma^2)$ for $i = 1,\dots,pun$. 
Since the convolution result is a vector of length $pun + l - 1$, the last $l - 1$ elements are discarded.
The variance of noise $\sigma^2 = (rE_b/N_0)^{-1}$, where the communication rate $r$ and energy per bit $E_b$ are defined by transmitter design, and the noise spectral density $N_0$ is a property of the channel. $N_0$ is accounted as a hyperparameter and remains constant during training.

\section{Simulation}
\label{sec:sim}

In order to estimate the performance of the AE for drill string communications, its BER, PAPR and power spectral density (PSD) were evaluated under a drill string channel. For comparison, an non-contiguous OFDM (NC-OFDM) was taken as a baseline. 

\subsection{Setup}

The simulation was carried out with complex signals at a sample rate $f_s = 2048$ Hz (Table \ref{tab:simparams}). The AE and baseline NC-OFDM  operate in a frequency band which starts at $f_c = 848$~Hz and has a width of $f_s / u = 512$~Hz.
Both were configured to have equal communication rate $r = m/n = 1.125$, which leads to the total throughput of 576 bits/s, given the specified bandwidth. 
But the latency of the AE system is $9$ times lower, since its symbol length is $64/(512+64) = 1/9$ of the NC-OFDM symbol length. In both cases $p$ is chosen so as to give equal packet length of $pun = 36 \cdot 4 \cdot 64  = 4  \cdot 4 \cdot (512+64) = 9216$ samples.
The channel impulse response was modeled using transfer matrix method (Sec.~\ref{sec:channel}) for a drill string composed of 10 pipes and 9 joints (Table~\ref{tab:drillparams}). 

\begin{table}[h]
    \caption{Parameters of simulation.}
    \label{tab:simparams}
    \begin{tabular}{rccccccc}
        \toprule
          & $m$ & $n$ & $p$ & $u$ & $l$ & $f_c$ (Hz) & $f_s$ (Hz)  \\
         \midrule
         AE & 72 &  64 & 36 & \multirow{2}{*}{4} & \multirow{2}{*}{2048} & \multirow{2}{*}{848} & \multirow{2}{*}{2048} \\ 
         NC-OFDM & 648 & 512 + 64 & 4 &  &  &  \\ 
         \bottomrule
    \end{tabular} 
\end{table}

\begin{table}[h]
    \caption{Properties of pipes (p) and joints (j).}
    \label{tab:drillparams}
    \begin{tabular}{cccccc}
        \toprule
        $d_p$ (mm) & $d_j$ (mm)&$a_p$ (cm$^2$)& $a_j$ (cm$^2$) & $c$ (m/s) & $\rho$ (kg/m$^3$) \\
         \midrule
         8760 & 240 & 52.276 & 248.186 & $ 5.13 \cdot 10^3$  & $7.87 \cdot 10^3$ \\
         \bottomrule
    \end{tabular} 
\end{table}

The baseline NC-OFDM system is a simplified version of a system proposed in \cite{Ma2018}. The PAPR reduction and pilot based channel estimation were omitted. Channel state information was assumed to be known at the receiver side. For the baseline NC-OFDM system $n$ denotes the length of OFDM symbol with CP, while $m$ is the number of bits carried by OFDM symbol. It uses QPSK bit mapping over 324 of 512 subcarries, which were non-contiguously selected within the two channel passbands ([878, 1049] Hz and [1169, 1320] Hz). 
The length of CP is 64 complex samples or 0.125~s after upsampling. The shorter CP leads to noticeable BER performance degradation.

The AE was trained for 256 epochs using Adam optimizer with initial learning rate 0.001. The training and test datasets contain 192 and 64 minibatches, respectively. Each minibatch consists of 128 packets, i.e., $128pm = 331776$~bits. For performance evaluation a distinct dataset of 256 minibatches was used for both the AE and the baseline NC-OFDM. All the data were sampled randomly from uniform distribution.

\subsection{Results}

The performance of the AE depends on the $N_0$ and $\alpha$ taken for training. 
For BER evaluation $\alpha$ was set to 0. In this case the AE ignores PAPR and tries to reconstruct input bit sequence as accurate as possible. PAPR reduction was investigated for $\alpha$ = 0.001 and 0.005. For each $\alpha$ the AE was trained at $1/N_0$ = 3 dB, 5 dB and 7 dB.

\subsubsection{Power Spectral Density}
\begin{figure}[h!]
    \centering
    \includegraphics[width=0.95\textwidth]{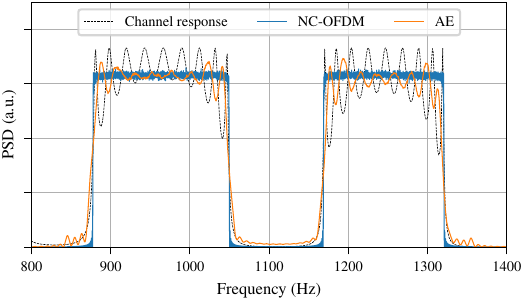}
    \caption{PSD of signal for the NC-OFDM and the AE trained at $1/N_0 = 3$~dB and $\alpha = 0$.}
    \label{fig:psd}
\end{figure}

While the PSD of the NC-OFDM signal is roughly constant within selected subbands, the PSD of the AE signal varies in accordance to the channel frequency response (Fig. \ref{fig:psd}). It illustrates how the AE adapts to the channel. 

\subsubsection{BER Performance}

\begin{figure}[h!]
    \centering
    \includegraphics[width=0.85\textwidth]{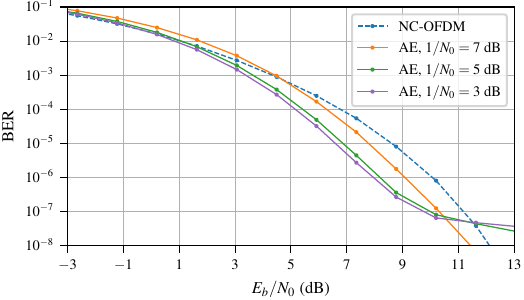}
    \caption{BER performance of the baseline NC-OFDM and the AE trained at different $N_0$ and $\alpha=0$.}
    \label{fig:ber}
\end{figure}

The AE is able to outperform the baseline in terms of BER for wide $E_b/N_0$ range (Fig. \ref{fig:ber}).  
AE trained at $1/N_0$ = 3 dB or 5 dB perform better for mid-range $E_b/N_0$, but yet suffers from ISI at higher $E_b/N_0$, so there is a BER floor. At high $N_0$ AWGN causes the majority of errors, therefore AE learns a signal that robust to AWGN and pays less attention to ISI.  Conversely, AE trained $1/N_0$ = 7 dB is more robust to ISI at high $E_b/N_0$, but performs worse at lower $E_b/N_0$. For low $N_0$ ISI causes the majority of errors, and the situation is symmetrically similar.



\subsubsection{PAPR Performance}







\begin{figure}[h!]
    \centering
    \includegraphics[width=0.95\textwidth]{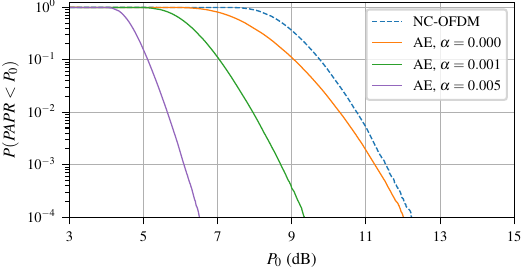}
    \caption{PAPR CCDF for the baseline NC-OFDM and the AE trained at $1/ N_0 = 3$~dB and different $\alpha$.}
    \label{fig:papr}
\end{figure}

\begin{figure}[h!]
    \centering
    \includegraphics[width=0.95\textwidth]{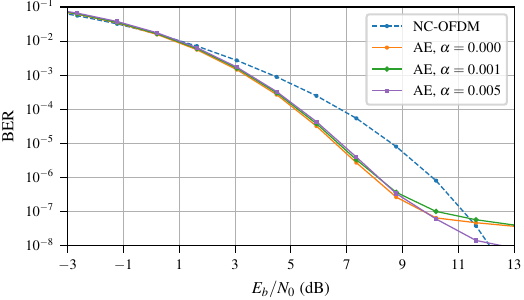}
    \caption{BER performance of the baseline NC-OFDM and the AE trained at $1/ N_0 = 3$~dB with different $\alpha$.}
    \label{fig:ber_papr}
\end{figure}

PAPR minimization during training results in a PAPR reduction (Fig. \ref{fig:papr}), but leads to a little degradation of BER performance (Fig. \ref{fig:ber_papr}). 
For $\alpha = 0.005$ the system is more robust at low $E_b/N_0$. In this case PAPR loss acts as a regularization, it helps AE to learn to perform better in a wider $E_b/N_0$ range. 


\section{Conclusion}

Simulation shows that AE is a perspective approach for acoustic drill string communications.
In comparison with the baseline NC-OFDM the AE is able to provide better BER and PAPR, operating at the same throughput and 9 times lower latency.
The AE adapts to the drill string channel during training and, thus, outperforms the baseline NC-OFDM, which is adapted manually by selecting non-contiguous carriers. 

However, AE has poor scalability in contrast to NC-OFDM. Increasing of transmission bandwidth or symbol duration, i.e., larger $m$ and $n$, may require a larger AE structure. As a result the curse of dimensionality is arisen. However, it is still possible but less efficient to use a batch of small AE systems multiplexed in frequency.

\backmatter











\bibliography{sn-bibliography}


\end{document}